\documentclass[letterpaper]{sig-alternate-2013}
\newfont{\mycrnotice}{ptmr8t at 7pt}
\newfont{\myconfname}{ptmri8t at 7pt}
\let\crnotice\mycrnotice%
\let\confname\myconfname%

\permission{Permission to make digital or hard copies of all or part of this work for personal or classroom use is granted without fee provided that copies are not made or distributed for profit or commercial advantage and that copies bear this notice and the full citation on the first page. Copyrights for components of this work owned by others than the author(s) must be honored. Abstracting with credit is permitted. To copy otherwise, or republish, to post on servers or to redistribute to lists, requires prior specific permission and/or a fee. Request permissions from Permissions@acm.org.}
\conferenceinfo{COSN'13,}{October 07--08, 2013, Boston, MA, USA. \\
{\mycrnotice{Copyright is held by the owner/author(s). Publication rights licensed to ACM.}}}
\copyrightetc{ACM \the\acmcopyr}
\crdata{978-1-4503-2084-9/13/10\ ...\$15.00.\\
http://dx.doi.org/10.1145/2512938.2512951}

\usepackage{graphicx}
\usepackage{url}
\usepackage{subfigure}
\usepackage{amsmath,amssymb,amsfonts}
\usepackage{cite}
\usepackage[T1]{fontenc}
\usepackage[english]{babel}
\usepackage{color}

\begin{document}

\title{Comparing and Combining Sentiment Analysis Methods}

\numberofauthors{4} 

\author{
\alignauthor Pollyanna Gon\c{c}alves\\
       \affaddr{UFMG}\\
       \affaddr{Belo Horizonte, Brazil}\\
       \email{pollyannaog@dcc.ufmg.br}
\alignauthor ~~~~~~Matheus Ara\'ujo~~~~~~\\
       \affaddr{UFMG}\\
       \affaddr{Belo Horizonte, Brazil}\\
       \email{matheus.araujo@dcc.ufmg.br}
\and
\alignauthor Fabr\'icio Benevenuto\\
       \affaddr{UFMG}\\
       \affaddr{Belo Horizonte, Brazil}\\
       \email{fabricio@dcc.ufmg.br}
\alignauthor Meeyoung Cha\\
       \affaddr{KAIST}\\
       \affaddr{Daejeon, Korea}\\
       \email{meeyoungcha@kaist.edu}
}

\maketitle

\begin{abstract}
Several messages express opinions about events, products, and services, political views or even their author's emotional state and mood. Sentiment analysis has been used in several
applications including analysis of the repercussions of events in social networks, analysis of opinions about products and services, and simply to better understand aspects of
social communication in Online Social Networks (OSNs).  There are multiple methods for measuring sentiments, including lexical-based approaches and supervised machine learning methods.  Despite the wide use and popularity of some methods, it is unclear which method is better for identifying the polarity (i.e., positive or negative) of a message as the current literature does not provide a method of comparison among existing methods. Such a comparison is crucial for understanding the potential limitations, advantages, and disadvantages of popular methods in analyzing the content of OSNs messages. Our study aims at filling this gap by presenting comparisons of eight popular sentiment analysis methods in terms of coverage (i.e., the fraction of messages whose sentiment is identified) and agreement (i.e., the fraction of identified sentiments that are in tune with ground truth). We develop a new method that combines existing approaches, providing the best coverage results and competitive agreement. We also present a free Web service called iFeel, which provides an open API for accessing and comparing results across different sentiment methods for a given text.

\end{abstract}


\section{Introduction}

Online Social Networks (OSNs) have become popular communication platforms for the public to logs thoughts, opinions, and sentiments about everything from social events to daily chatter.  The size of the active user bases and the volume of data created daily on OSNs are massive. Twitter, a popular micro-blogging site, has 200 million active users, who post more than 400 million tweets a day~\cite{twitter}. Notably, a large fraction of OSN users make their content public (e.g., 90\% in case of Twitter), allowing
researchers and companies to gather and analyze data at scale~\cite{cha_icwsm10}. As a result, a big number of studies have monitored the trending topics, memes, and notable events on OSNs, including political events~\cite{Tumasjan}, stock marketing  fluctuations~\cite{DBLP:journals/corr/abs-1010-3003}, disease
epidemics~\cite{gomide2010dengue,lamb-paul-dredze:2013:NAACL-HLT}, and natural disasters~\cite{Sakaki@www10}.

One important tool used in this context is methods for detecting sentiments expressed in OSN messages. While a wide range of human moods can be captured through sentiment analysis, a large majority of studies focus on identifying the \textit{polarity} of a given text---that is to automatically identify if a message about a certain topic is positive or negative. Polarity analysis has numerous applications especially for real time systems that rely on analyzing public opinions  or mood fluctuations (e.g., social network analytics on product launches)~\cite{hannak-2012-weather}.

Broadly, there exist two types of methods for sentiment analysis: machine-learning-based and lexical-based. Machine learning methods often rely on supervised classification approaches, where sentiment detection is framed as a binary (i.e., positive or negative). This approach requires labeled data to train classifiers~\cite{Pang:2002:TUS:1118693.1118704}. While one advantage of learning-based  methods is their ability to adapt and create trained models for specific purposes and contexts, their drawback is the availability of labeled data and hence the low applicability of the method on new data. This is because labeling data might be costly or even prohibitive for some tasks.

On the other hand, lexical-based methods make use of a predefined list of words, where each word is associated with a specific sentiment. The lexical methods vary according to the context in which they were created. For instance, LIWC~\cite{liwc} was originally proposed to analyze sentiment patterns in formally written English texts, whereas PANAS-t~\cite{polly@panast} and POMS-ex~\cite{Bollen} were proposed as psychometric scales adapted to the Web context. Although lexical methods do not rely on labeled data, it is hard to create a unique lexical-based dictionary to be used for different contexts. For instance, slang is common in OSNs but is rarely supported in lexical methods~\cite{WWW2013Hu}.

Despite business potentials, little is known about how various sentiment methods work in the context of OSNs. In practice, sentiment methods have been widely used for developing
applications without an understanding either of their applicability in the context of OSNs, or their advantages, disadvantages, and limitations in comparison with one another. In
fact, many of these methods were proposed for complete sentences, not for real-time short messages, yet little eff-ort has been paid to apple-to-apple comparison of the most widely
used sentiment analysis methods. The limited available research shows machine learning approaches (Na\"{i}ve Bayes, Maximum Entropy, and SVM) to be more suitable for Twitter than the
lexical-based LIWC method~\cite{liwc}. Similarly, classification methods (SVM, and Multinomial Na\"{i}ve Bayes) are more suitable than SentiWordNet for
Twitter~\cite{Bermingham:2010:CSM:1871437.1871741}. However, it is hard to conclude whether a single classification method is better than all lexical methods across different
scenarios nor if it can achieve the same level of coverage as some lexical methods.

In this paper, we aim to fill this research gap.  We use two different sets of OSN data to compare eight widely used sentiment analysis methods: LIWC, Happiness Index, SentiWordNet, SASA, PANAS-t, Emoticons, SenticNet, and SentiStrength. As a first step comparison, we focus on determining the polarity (i.e., positive and negative affects) of a given social media text, which is an overlapping dimension across all eight sentiment methods and provides desirable information for a number of different applications. The
two datasets we employ are large in scale. The first consists of about 1.8 billion Twitter messages~\cite{cha_icwsm10}, from which we present six major events, including tragedies, product releases, politics, health, and sports.  The second dataset is an extensive collection of texts, whose sentiments were labeled by humans~\cite{sentistrength1}. Based on these datasets, we compare the eight sentiment methods in terms of coverage (i.e., the fraction of messages whose sentiment is identified) and
agreement (i.e., the fraction of identified sentiments that are in tune with results from others).

We summarize some of our main results:
\begin{enumerate}
\item Existing sentiment analysis methods have varying degrees of coverage, ranging between 4\% and 95\% when applied to real events. This means that depending on the sentiment method used, only a small fraction of data may be analyzed, leading to a bias or under-representation of data.

\item  No single existing sentiment analysis method had high coverage and correspondingly high agreement. Emoticons achieve the highest agreement of above 85\%, but have extremely low agreement of between 4\% to 13\%.

\item  When it comes to the predicted polarity, existing methods varied widely in their agreement, ranging from 33\% to 80\%. This suggests that the same social media text could be interpreted very differently depending on the choice of a sentiment method.

\item  Existing methods varied widely in their sentiment prediction of notable social events. For the case of an airplane crash, half of the methods predicted the relevant tweets to contain positive affect, instead of negative affect. For the case of a disease outbreak, only two out of eight methods predicted the relevant tweets to contain negative affect.

\end{enumerate}

Finally, based on these observations, we developed a new sentiment analysis method that combines all eight existing approaches in order to provide the best coverage and competitive agreement. We further implement a public Web API, called iFeel (\url{http://www.ifeel.dcc.ufmg.br}), which provides comparative results among the different sentiment methods for a given text. We hope that our tool will help those researchers and companies interested in an open API for accessing and comparing a wide range of sentiment analysis techniques.

The rest of this paper is organized as follows. In Section~2, we describe the eight methods that are used for comparison, as we cover a wide set of related work. Section~3 outlines the comparison methodology as well as the data used for comparison, and Section~4 highlights the comparison results. In Section~5, we propose a newly combined method of sentiment analysis that has the highest coverage in handling OSN data, while having reasonable agreement. We present the iFeel system and conclude in Section~6.

\section{Sentiment Analysis Methods}

This section provides a brief description of the eight sentiment analysis methods investigated in this paper. These methods are the most popular in the literature (i.e., the most cited and widely used) and they cover diverse techniques such as the use of Natural Language Processing (NLP) in assigning polarity, the use of Amazon's Mechanical Turk (AMT) to create labeled datasets, the use of psychometric scales to identify mood-based sentiments, the use of supervised and
unsupervised machine learning techniques, and so on. Validation of these methods also varies greatly, from using toy examples to a large collection of labeled data.

\subsection{Emoticons}

The simplest to detect the way polarity (i.e., positive and negative affect) of a message is based on the emoticons it contains. Emoticons have become popular in recent years, to the extent that some (e.g. \verb1<31) are now included in English Oxford Dictionary~\cite{oxford}. Emoticons are primarily face-based and represent happy or sad feelings, although a wide range of non-facial variations exist: for instance, \verb1<31 represents a heart and expresses love or affection.

To extract polarity from emoticons, we utilize a set of common emoticons from~\cite{emoticons, msn, yahoo} as listed in Table~\ref{tab:emoticons}.  This table also includes the popular variations that express the primary polarities of positive, negative, and neutral. Messages with more than one emoticon were associated to the polarity of the first emoticon that appeared in the text, although we encountered only  a small number of such cases in the data.

\begin{table}[htpb!]
	\caption{Emoticons and their variations}
		\small
		\begin{tabular}{| p{1.4cm} | p{1.2cm}  | p{4.7cm} |}
				\hline
				 \textbf{Emoticon} &  \textbf{Polarity} &  \textbf{Symbols}\\
				\hline
				& &\\
				& & \verb1 :)  :]  :}  :o)  :o]  :o} 1 \\
				& & \verb1 :-] :-)  :-}  =)  =]  =}1\\
				& & \verb1 =^]  =^)  =^}  :B  :-D  :-B 1 \\
				\centering \includegraphics[scale=0.5]{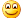} &  Positive & \verb1  :^D  :^B  =B  =^B  =^D  :') 1 \\
				& & \verb1 :']  :'}  =')  =']  ='}  <3 1\\
				& & \verb1 ^.^  ^-^  ^_^  ^^  :*  =* 1 \\
				& & \verb1 :-*  ;)  ;]  ;}  :-p  :-P 1\\
				& & \verb1 :-b  :^p  :^P  :^b   =P  1 \\
				& & \verb1 =p  \o\  /o/ :P :p  :b  =b  1 \\
				& & \verb1 =^p =^P  =^b \o/ 1 \\ \hline
				& &\\
				& & \verb1 D:  D=  D-:  D^:  D^=  :(  :[ 1 \\
				& & \verb1 :{  :o(   :o[   :^(  :^[  :^{  1\\
				& & \verb1 =^(  =^{  >=(  >=[  >={  >=( 1 \\
				& & \verb1 >:-{  >:-[  >:-(  >=^[  >:-( 1 \\
				\centering \includegraphics[scale=0.5]{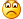} &  Negative & \verb1 :-[  :-(  =(  =[  ={  =^[  1 \\
				& & \verb1 >:-=(  >=[  >=^(  :'(  :'[ 1\\
				& & \verb1 :'{  ='{  ='(  ='[  =\  :\ 1 \\
				& & \verb1  =/  :/  =$  o.O  O_o  Oo 1\\
				& & \verb1 :$:-{  >:-{  >=^{  :o{  1\\ \hline
				& &\\
				& & \verb1 :|  =|  :-|  >.<  ><  >_<  :o 1 \\
				\centering \includegraphics[scale=0.5]{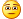} &  Neutral & \verb1 :0  =O  :@  =@  :^o  :^@  -.- 1 \\
				& & \verb1 -.-'  -_-  -_-'  :x  =X  :#  1 \\
				& & \verb1 =#  :-x  :-@  :-#  :^x  :^# 1\\ \hline				
	\end{tabular}
	\label{tab:emoticons}
\end{table}

As one may expect, the rate of OSN messages containing at least one emoticon is very low compared to the total number of messages that could express emotion. A recent work has identified that this rate is less than 10\%~\cite{park_icwsm13}. Therefore,  emoticons have been often used
in combination with other techniques for building  a training dataset in supervised machine learning techniques~\cite{Read-emoticons-for-training}.

\subsection{LIWC}

LIWC (Linguistic Inquiry and Word Count)~\cite{liwc} is a  text analysis  tool that evaluates emotional, cognitive, and structural components of a given text based on the use of a dictionary containing words and their classified categories. In addition to detecting positive and negative affects in a given text, LIWC provides other sets of sentiment categories. For example, the word ``agree" belongs to the following word categories: assent, affective, positive emotion, positive feeling, and cognitive process.

The LIWC software is commercial and provides optimization options such as allowing users to include customized dictionaries instead of the standard ones. For this paper, we used the LIWC2007 version and its English dictionary, which is the most current version and contains labels for more than 4,500 words and 100 word categories. The LIWC software is available at \url{http://www.liwc.net/}. In order to measure polarity, we examined the relative rate of positive and negative affects in the feeling categories.

\subsection{SentiStrength}

Machine-learning-based methods are suitable for applications that need content-driven or adaptive polarity identification models. Several key classifiers for identifying polarity in OSN data have been proposed in the literature~\cite{Bermingham:2010:CSM:1871437.1871741,Paltoglou:2012,sentistrength1}.

The most comprehensive work~\cite{sentistrength1} compared a wide range of supervised and unsupervised classification methods, including simple logistic regression, SVM, J48 classification tree, JRip rule-based classifier, SVM regression, AdaBoost, Decision Table, Multilayer Perception, and Na\"{i}ve Bayes. The core classification of this work relies on the set of words in the LIWC dictionary~\cite{liwc}, and the authors expanded this baseline by adding new  features for the OSN context. The features added include a list of negative and positive words, a list of booster words to strengthen (e.g., ``very'') or weaken (e.g., ``somewhat'') sentiments, a list of emoticons, and the use of repeated punctuation (e.g., ``Cool!!!!'') to strengthen sentiments.  For evaluation, the authors used labeled text messages from six different Web 2.0 sources, including MySpace, Twitter, Digg, BBC Forum, Runners World Forum, and YouTube Comments.

The authors released a tool named SentiStrengh, which implements a combination of learning techniques that produces the best results and the best training model
empirically obtained~\cite{sentistrength1}. Therefore, SentiStrengh implements the state-of-the-art machine learning method in the context of OSNs. We used SentiStrength version 2.0, which
is available at \url{http://sentistrength.wlv.ac.uk/Download}.

\subsection{SentiWordNet}

SentiWordNet~\cite{sentiwordnet} is a tool that is widely used in opinion mining, and is based on an English lexical dictionary called WordNet~\cite{wordnet}. This lexical dictionary groups adjectives, nouns, verbs and other grammatical classes into synonym sets called synsets. SentiWordNet associates three scores with synset from the WordNet dictionary to indicate the sentiment of the text: positive, negative, and objective (neutral). The scores, which are in the values of [0, 1] and add up to 1, are obtained using a semi-supervised machine learning method. For example, suppose that a given synset ${s} = [bad, wicked, terrible]$ has been extracted from a tweet. SentiWordNet then will give scores of 0.0 for positive, 0.850 for negative, and 0.150 for objective sentiments, respectively. SentiWordNet was evaluated with a labeled lexicon dictionary.

In this paper, we used SentiWordNet version 3.0, which is available at \url{http://sentiwordnet.isti.cnr.it/}.  To assign polarity based on this method, we considered the average scores of all associated synsets of a given text and consider it to be positive, if the average score of the positive affect is greater than that of the negative affect. Scores from objective sentiment were not used in determining polarity.

\subsection{SenticNet}

SenticNet~\cite{FSS102216} is a method of opinion mining and sentiment analysis that explores artificial intelligence and semantic Web techniques. The goal of SenticNet is to infer the polarity of common sense concepts from natural language text at a semantic level, rather than at the syntactic level. The method uses Natural Language Processing (NLP) techniques to create a polarity for nearly 14,000 concepts. For instance, to interpret a message ``Boring, it's Monday morning", SenticNet first tries to identify concepts, which are ``boring" and ``Monday morning" in this case. Then it gives polarity score to each concept, in this case, -0.383 for ``boring", and +0.228 for ``Monday morning". The resulting sentiment score of SenticNet for this example is -0.077, which is the average of these values.

SenticNet was tested and evaluated as a tool to measure the level of polarity in opinions of patients about the National Health Service in England~\cite{cambria@websci2010}. The authors also tested SenticNet with data from LiveJournal blogs, where posts were labeled by the authors with over 130 moods, then categorized as either positive or negative~\cite{Read-emoticons-for-training, Somasundaran@coling08}. We use SenticNet version 2.0, which is available at \url{http://sentic.net/}.

\subsection{SASA}

We employ one more machine learning-based tool called the Sail\/Ail Sentiment Analyzer (SASA)~\cite{sasa}.  SASA is a method based on machine learning techniques such as SentiStrengh and was evaluated with 17,000 labeled tweets  on the 2012 U.S. Elections. The open source tool was evaluated by the Amazon Mechanical Turk (AMT)~\cite{Turkey}, where ``turkers'' were invited to label tweets as positive, negative, neutral, or undefined.  We include SASA in particular because it is an open source tool and further because there had been no apple-to-apple comparison of this tool against other methods in the sentiment analysis literature.  We used the SASA python package version
0.1.3, which is available at \url{https://pypi.python.org/pypi/sasa/0.1.3}.

\begin{table*}[htpb!] \centering \caption{Summary information of the
six major topics events studied}\label{tab:events}
\vspace*{2mm}
\begin{tabular} {|l|l|l|}
\hline
Topic   &Period  &Keywords  \cr \hline
\textsf{AirFrance} &06.01--06.2009     &victims, passengers, a330, 447, crash, airplane, airfrance  \cr
\textsf{2008US-Elect}     &11.02--06.2008  &voting, vote, candidate, campaign, mccain, democrat*, republican*, obama, bush   \cr
\textsf{2008Olympics}     &08.06--26.2008  &olympics, medal*, china, beijing, sports, peking, sponsor   \cr
\textsf{Susan Boyle}     &04.11--16.2009  &susan boyle, I dreamed a dream, britain's got talent, les miserables  \cr
\textsf{H1N1}     &06.09--26.2009 &outbreak, virus, influenza, pandemi*, h1n1, swine, world health organization \cr
\textsf{Harry-Potter}   &07.13--17.2009  &harry potter, half-blood prince, rowling  \cr \hline
\end{tabular} 
\end{table*}

\subsection{Happiness Index}

Happiness Index~\cite{DoddsP2009Measuring} is a sentiment scale that uses the popular Affective Norms for English Words (ANEW)~\cite{citeulike:3519108}.  ANEW is a collection of 1,034 words commonly used associated with their affective dimensions of valence, arousal, and dominance.  Happiness Index was constructed based on the ANEW terms and has scores for a given text between 1 and 9, indicating the amount of happiness existing in the text. The authors calculated the frequency that each word from the ANEW appears in the text and then computed a weighted average of the valence of the ANEW study words. The validation of the Happiness Index score is based on examples. In particular, the authors applied it to a dataset of song lyrics, song titles, and blog sentences. They found that the happiness score for song lyrics had declined from 1961 to 2007, while the score for blog posts in the same period had increased.

In order to adapt Happiness Index for detecting polarity, we considered any text that is classified with this method in the range of $[1..5)$ to be negative and in the range of $[5..9]$) to be positive.

\subsection{PANAS-t}

The PANAS-t~\cite{polly@panast} is a psychometric scale proposed by us for detecting mood fluctuations of users on Twitter. The method consists of an adapted version of the {Positive Affect Negative Affect Scale} (PANAS)~\cite{Watson}, which is a well-known method in psychology. The PANAS-t is based on a large set of words associated with eleven moods: joviality, assurance, serenity, surprise, fear, sadness, guilt, hostility, shyness, fatigue, and attentiveness. The method is designed to track any increase or decrease in sentiments over time.

To associate text to a specific sentiment, PANAS-t first utilizes a baseline or the normative values of each sentiment based on the entire data. Then the method computes the ${P(s)}$ score for each sentiment ${s}$ for a given time period as values between  $[-1.0, 1.0]$ to indicate the change. For example, if a given set of tweets contain ${P(``surprise")}$ as 0.250, then sentiments related to ``surprise'' increased by 25\% compared to a typical day.  Similarly, ${P(s)} = -0.015$ means that the sentiment $s$ decreased by 1.5\% compared to a typical day. For evaluation, we presented evidence that the method works for tweets about noteworthy events. In this paper, we consider joviality, assurance, serenity, and surprise to be positive affect and fear, sadness, guilt, hostility, shyness, and fatigue to be negative affect. We consider attentiveness to be neutral.

Another method similar to PANAS-t is an adaptation of the Profile of Mood States (POMS)~\cite{Bollen}, a psychological rating scale that measures certain mood states consisting of 65 adjectives that qualify the following feelings: tension, depression, anger, vigor, fatigue and confusion. However, we could not include this method for comparison as it was not made publicly available upon request.

\section{Methodology}

Having introduced the eight sentiment analysis methods, we now describe the datasets and metrics used for comparison.

\subsection{Datasets}

We employ two different datasets in this paper.

\subsubsection{Near-complete Twitter logs}

The first dataset is a near-complete log of Twitter messages posted by all users from March 2006 to August 2009~\cite{cha_icwsm10}. This dataset contains 54 million users who had 1.9 billion follow links among themselves and posted 1.7 billion tweets over the course of 3.5 years. This dataset is appropriate for the purpose of this work as it contains all users who set their account publicly available (excluding those users who set their accounts private) and their tweets, which is not based on sampling and hence alleviates any sampling bias. Additionally, this dataset allows us to study the reactions to noteworthy past events and evaluate our methods on data from real scenarios.

We chose six events covered by Twitter users\footnote{Top Twitter trends at \url{http://tinyurl.com/yb4965e}}. These events, summarized in Table~\ref{tab:events}, span topics related to tragedies, product and movie releases, politics, health and sports events. To extract tweets relevant to these events, we first identified the sets of keywords describing the topics by consulting news websites, blogs, Wikipedia, and informed individuals. Given our selected list of keywords, we identified the topics by searching for keywords in the tweet dataset. This process is very similar to the way in which  mining and monitoring tools to crawl data about specific topics.

We limited the duration of each event because popular keywords are typically hijacked by spammers after a certain amount of time. Table~\ref{tab:events} displays the keywords used and the total number of tweets for each topic. The first column contains a short name for the event, which we use to refer to them in the rest of the paper. While the table does not show the ground truth sentiment of the six events, we can utilize these events to compare the predicted sentiments across different methods.

\subsubsection{Labeled Web 2.0 data}

The second dataset is six sets of messages labeled as positive and negative by humans, and was made available in the SentiStrength research~\cite{sentistrength1}.  These datasets include a wide range of social web texts from: MySpace, Twitter, Digg, BBC forum, Runners World forum, and YouTube comments. Table~\ref{tab:sentistrenght_corpus} summarizes the number of messages in each dataset along with the fraction of positive and negative ground truth.

\begin{table}[!htb]
            \caption{Labeled datasets}
            \centering
                \begin{tabular}{| l | c | c |}
                \hline
                \textbf{Data type} & \textbf{\# Messages} & \textbf{Pos / Neg}\\
                \hline
                Twitter& 4,242 & 58.58\% / 41.42\% \\ 
                MySpace& 1,041 & 84.17\% / 15.83\%  \\ 
                YouTube & 3,407 & 68.44\% / 31.56\% \\  
                BBC forum & 1,000 & 13.16\% / 86.84\%  \\ 
                Runners world& 1,046 & 68.65\% / 31.35\% \\  
                Digg  & 1,077 & 26.85\% / 73.15\% \\  \hline
                \end{tabular}
                \label{tab:sentistrenght_corpus}
\end{table}

With this human-labeled data, we are able to quantify the extent to which different sentiment analysis methods can accurately predict polarity of content. We do not measure this for SentiStrength, since this method is trained on the same dataset.

\begin{figure*}[htpb!]
  \centering {
    \subfigure[AirFrance]{\includegraphics[width=0.32\textwidth]{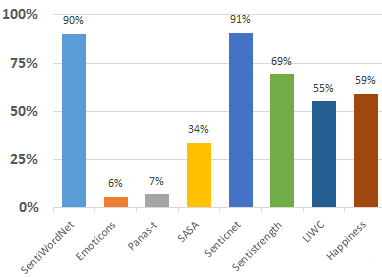}}
		 \subfigure[2008Olympics]{\includegraphics[width=0.32\textwidth]{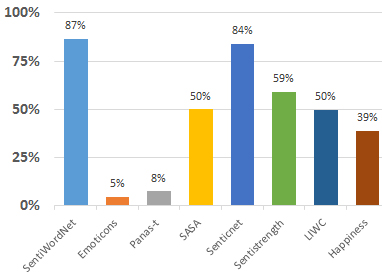}}
		\subfigure[Susan Boyle]{\includegraphics[width=0.32\textwidth]{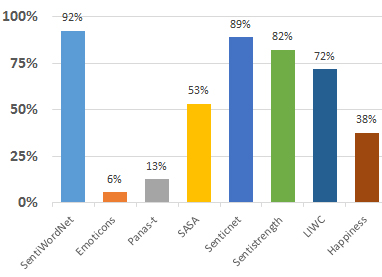}}
		\subfigure[US-Elect]{\includegraphics[width=0.32\textwidth]{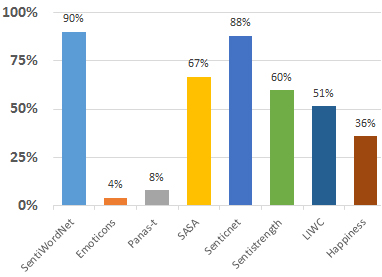}}
		\subfigure[H1N1]{\includegraphics[width=0.32\textwidth]{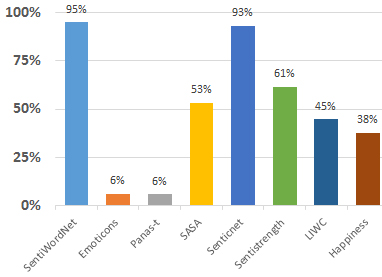}}
		\subfigure[Harry Potter]{\includegraphics[width=0.32\textwidth]{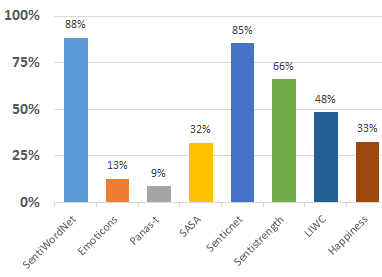}}
  }
  \caption{Coverage of six events.}\label{fig:coverage_all}
\end{figure*}

\subsection{Comparison Measures}

In order to define the metrics used to evaluate the methods we are analyzing, we consider the following metrics:

\begin{table}[h] \centering
    \begin{tabular}{|c|c|cc|} \hline
      \multicolumn{2}{|c}{} & \multicolumn{2}{|c|}{\textit{Actual observation}} \\
     \multicolumn{2}{|c|}{}   & Positive & Negative \\ \hline
    \textit{Predicted} & Positive & a        & b        \\
    \textit{~~expectation~~} & Negative & c        & d        \\ \hline
    \end{tabular}
\end{table}

Let \textit{a} represent the number of messages correctly classified as positive (i.e., true positive), \textit{b} the number of negative messages classified as positive (i.e., false positive), \textit{c} the number of positive messages classified as negative (i.e., false negative), and \textit{d} the number of messages correctly classified as negative (i.e., true negative). In order to compare and evaluate the methods, we consider the following metrics, commonly used in information retrieval: true positive rate or recall: $R = a/(a+c)$, false positive rate or precision: $P = a/(a+b)$, accuracy: $A = (a+d)/(a+b+c+d)$, and F-measure: $F = 2\cdot (P\cdot R)/(P+R)$. We will in many cases simply use the F-measure, as it is a measure of a test's accuracy and relies on both precision and recall.

We report all the metrics listed above since they have direct interpretation in practice. The true positive rate or recall can be understood as the rate at which positive messages are predicted to be positive ($R$), whereas the true negative rate is the rate at which negative messages are predicted to be negative.  The accuracy represents the rate at which the method predicts results correctly ($A$). The precision rate, also called the positive predictive rate, calculates how close the measured values are to each other ($P$). We also use the F-measure to compare results, since it is a standard way of summarizing precision and recall ($F$). Ideally, a polarity identification method reaches the maximum value of the F-measure, which is 1, meaning that its polarity classification is perfect.

Finally, we define \textit{coverage} as the fraction of messages in a given dataset that a method is able to classify as either positive or negative. Ideally, polarity detection methods should retain high coverage to avoid bias in the results, due to the unidentified messages. For instance, suppose that a sentiment method has classified only 10\% of a given set of tweets. The remaining 90\% consisting of unidentified tweets may completely change the result, that is, whether the context drawn from tweets should be positive or negative. Therefore, having high coverage in data is essential in analyzing OSN data. In addition to high coverage, it is also desirable to have a high F-measure as discussed above.

\begin{table*}[t]
            \caption{Percentage of agreement between methods.}
            \centering
            \small
                \begin{tabular}{| l | c | c | c | c | c | c | c | c | c |}
                \hline
                \textbf{ } & \textbf{ }  & \textbf{ }  & \textbf{ }  & \textbf{Sentic-}  & \textbf{Senti-}  & \textbf{Happiness}  & \textbf{Senti-}   & \textbf{ } & \textbf{ } \\ 
                \textbf{Metric} & \textbf{PANAS-t}  & \textbf{Emoticons}  & \textbf{SASA}  & \textbf{Net}  & \textbf{WordNet}  & \textbf{Index}  & \textbf{Strength} & \textbf{LIWC} & \textbf{Average}\\ \hline
                \textbf{PANAS-t} & - & 60.00 & 66.67 & 30.77 & 56.25 & - & 74.07 & 80.00 & 52.53 \\ 
                \textbf{Emoticons} & 33.33 & - & 64.52 & 64.00 & 57.14 & 58.33 & 72.00 & 75.00 & 60.61 \\ 
                \textbf{SASA} & 66.67 & 64.52 & - & 64.29 & 60.00 & 64.29 & 61.76 & 68.75 & 64.32 \\ 
                \textbf{SenticNet} & 30.77 & 60.00 & 64.29 & - & 64.29 & 59.26 & 63.33 & 73.33 & 59.32\\ 
                \textbf{SentiWordNet} & 56.25 & 57.14 & 60.00 & 64.29 & - & 64.10 & 52.94 & 62.50 & 59.04 \\ 
                \textbf{Happiness Index} & - & 58.33 & 64.29 & 62.50 & 70.27 & - & 65.52 & 71.43 & 56.04\\ 
                \textbf{SentiStrength} & 74.07 & 75.00 & 63.89 & 63.33 & 52.94 & 65.52 & - & 75.00 & 66.67\\ 
                \textbf{LIWC} & 80.00 & 75.00 & 68.97 & 73.33 & 58.82 & 83.33 & 75.00 & - & 73.49 \\ \hline
                                \textbf{Average} & 48.72 & 63.85 & 64.65 & 60.35 & 59.95 & 56.40 & 66.37 & 72.29 & - \\ \hline

                \end{tabular}
                \label{tab:agreement_methods}
\end{table*}

\begin{table*}[t]
			\caption{Average prediction performance for all labeled dataset.}			
			\centering
				\begin{tabular}{| l | c | c | c | c | c | c | c | c |}
				\hline
				\textbf{ } & \textbf{ }  & \textbf{ }  & \textbf{ }  & \textbf{Sentic-}  & \textbf{Senti-}  & \textbf{Happiness}  & \textbf{Senti-}   & \textbf{}\\
				\textbf{Metric} & \textbf{PANAS-t}  & \textbf{Emoticons}  & \textbf{SASA}  & \textbf{Net}  & \textbf{WordNet}  & \textbf{Index}  & \textbf{Strength}   & \textbf{LIWC}\\
				\hline
				\textbf{Recall} & 0.614 & 0.856 & 0.648 & 0.562 & 0.601 & 0.571 & 0.767  & 0.153 \\
				\textbf{Precision} & 0.741 & 0.867 & 0.667 & 0.934 & 0.786 & 0.945 & 0.780 & 0.846\\
				\textbf{Accuracy} & 0.677 & 0.817 & 0.649 & 0.590 & 0.643 & 0.639 & 0.815 & 0.675\\
				\textbf{F-measure} & 0.632 & 0.846 & 0.627 & 0.658 & 0.646 & 0.665 & 0.765 & 0.689\\ \hline
				\end{tabular}
				\label{tab:metrics_all}
\end{table*}

\section{Comparison Results}

In order to understand the advantages, disadvantages, and limitations of the various sentiment analysis methods, we present comparison results among them.

\subsection{Coverage}

We begin by comparing the coverage of all methods across the representative events from Twitter and also examine the intersection of the covered tweets across the methods.

For each topic described in Table~\ref{tab:events}, we computed the coverage of each of the eight sentiment analysis methods. Figure~\ref{fig:coverage_all}(a) shows the result for the AirFrance event, a tragic plane crash in 2009. As shown in the figure, SentiWordNet and SenticNet have the highest coverage with 90\% and 91\%, respectively, followed by SentiStrength with 69\%. Emoticons and PANAS-t can interpret less than 10\% of all relevant tweets. In the case of the U.S. Election event depicted in Figure~\ref{fig:coverage_all}(d), SentiWordNet, SenticNet and SASA have the highest coverage percentages with 90\%, 88\% and 67\%, respectively. 

In fact, either SentiWordNet and SenticNet had the highest coverage for every event from Table~\ref{tab:events}. In the other events SentiStrength, LIWC, and SASA had ranked in third and fourth positions.

Figure~\ref{fig:coverage_all}(e) shows the result for the outbreak of the H1N1 influenza, a worldwide epidemic declared by the World Health Organization in 2009. In this case, SentiWordNet and SenticNet have the highest coverage with 95\% and 93\%, respectively, followed by SentiStrength with 61\%. The ranking of coverage across the methods is similar to the AirFrance event.

The analysis above shows that despite a few methods having high coverage, the percentage of tweets left unidentified is significant for most of methods, which is a serious problem for sentiment analysis. We next examine what fraction of the tweets can be identified if we combine more than one method. For each event, we combined all methods one by one, beginning from the one with the highest coverage to the one with the lowest coverage. Combining two methods, we were able to increase the coverage to more than 92.75\% for each of the events.  We also noted that using this strategy the percentage of uncovered tweets is smaller than 7.24\% for each of the events. This result is important as we will shortly demonstrate that combining  methods can increase the coverage over a single method.

\subsection{Agreement}

Next we examine the degree to which different methods agree on the polarity of the content. For instance, when two or more methods detect sentiments in the same message it is important to check whether these sentiments are the same; this would strengthen the confidence in the polarity classification. In order to compute the agreement of each method, we calculated the intersections of the positive or negative proportion given by each method.

Table~\ref{tab:agreement_methods} presents the percentage of agreement for each method with all the others. For each method in the first column, we measure, from the messages classified for each pair of methods, for what fraction of these messages they agree. We find that some methods have a high degree of overlap as in the case of LIWC and PANAS-t (80\%), while others have very low overlap such as SenticNet and PANAS-t (30.77\%). PANAS-t and Happiness Index had no intersection. The last ``column'' of the table shows on average to what extent each method agrees with the other seven, whereas the last ``row'' quantifies how other methods agree with a certain method, on average. In both situations, the method that most agrees with others and which others agree with it is LIWC, suggesting that LIWC might provide an interesting method to be used in combination with others.

In summary, the above result indicates that existing tools vary widely in terms of agreement about the predicted polarity, with scores ranging from 33\% to 80\%. This implies that the same social media text, when analyzed with different sentiment tools, could be interpreted very differently. In particular, for those tools that have lower than 50\% agreement, the polarity will even change (e.g., from positive to negative, or vice versa).

\begin{table*}[t]
			\caption{F-measures for the eight methods.}			
			\centering
				\begin{tabular}{| l | c | c |c | c |c |c |}
				\hline
				\textbf{Method} & \textbf{Twitter} & \textbf{MySpace} & \textbf{YouTube} & \textbf{BBC} & \textbf{Digg} & \textbf{Runners World}\\
				\hline
				PANAS-t & 0.643 & 0.958 & 0.737 & 0.296 & 0.476 & 0.689 \\ 
				Emoticons & 0.929 & 0.952 & 0.948 & 0.359 & 0.939 & 0.947 \\ 
				SASA & 0.750 & 0.710 & 0.754 & 0.346 & 0.502 & 0.744 \\ 
				SenticNet & 0.757 & 0.884 & 0.810 & 0.251 & 0.424 & 0.826\\ 
				SentiWordNet & 0.721 & 0.837 & 0.789 & 0.284 & 0.456 & 0.789 \\ 
				SentiStrength & 0.843 & 0.915 & 0.894 & 0.532 & 0.632 & 0.778 \\ 
				Happiness Index  & 0.774 & 0.925 & 0.821 & 0.246 & 0.393 & 0.832 \\ 
				LIWC & 0.690 & 0.862 & 0.731 & 0.377 & 0.585 & 0.895\\ \hline
				\end{tabular}
				\label{tab:fmeasures}
\end{table*}

\newpage
\subsection{Prediction Performance}

Next we present a comparative performance evaluation of each method in terms of correctly predicting polarity. Here we present the results for precision, recall, accuracy, and F-measure for the eight methods. To compute these metrics, we used the the ground truth provided by SentiStrength's dataset~\cite{sentistrength1}.

In order to compare the results of prediction performance for each method we present Table~\ref{tab:metrics_all}, which gives the average of the results obtained for each labeled dataset. For the F-measure, a score of 1 is ideal and 0 is the worst possible. The method with the best F-measure was Emoticons (0.846), which had the lowest coverage. The second best method in terms of F-measure is SentiStrength, which obtained a much higher coverage than Emoticons. It is important to note that the SentiStrength version we are using is already trained, probably with this entire dataset. Thus, running experiments with SentiStrength using this dataset would be potentially biased, as it would be training and testing with the same dataset. Instead, we compute the prediction performance metrics for SentiStrengh based on the numbers they reported in their experiments~\cite{sentistrength1}.

Table~\ref{tab:fmeasures} presents the F-measures calculated for each analysis method and each of the labeled datasets we are using. Overall, we note that the eight methods yielded wide variation in their results across the different datasets. We observe better performance on datasets that contain more expressed sentiment, such as social network messages (e.g., Twitter and MySpace) and lower performance on formal datasets (e.g., BBC and Digg). For instance, on BBC posts (i.e., formal content), the highest F-measure was 0.532, from SentiStrength
On the other hand, for the MySpace dataset (i.e., informal content), the highest F-measure was obtained by PANAS-t (0.958) and the average F-measure for all 8 methods was 72\%. This might indicate that each method complements the others in different ways.

\subsection{Polarity Analysis}

Thus far, we have analyzed the coverage prediction performance of the sentiment analysis methods. Next, we provide a deeper analysis on how polarity varies across different datasets and potential pitfalls to avoid when monitoring and measuring polarity.

\begin{figure*}[htpb!]
  \centering {
   \includegraphics[width=0.9\textwidth]{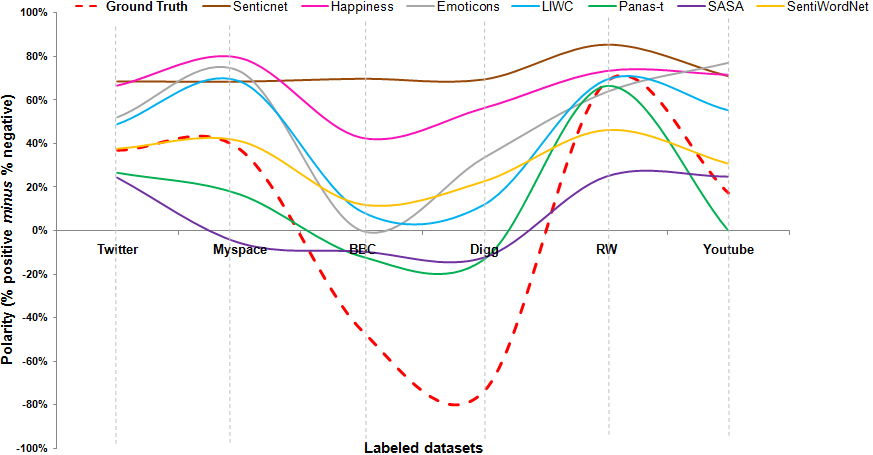}
  }
  \caption{Polarity of the eight sentiment methods across the labeled datasets, indicating that existing methods vary widely in their agreement.}\label{fig:polarity-labeled}
\end{figure*}

Figure~\ref{fig:polarity-labeled} presents the polarity of each method when exposed to each labeled dataset. For each dataset and method, we computed the percentage of positive messages and the percentage of negative messages. The Y-axis shows the positive percentage minus the negative percentage.  We also plot the ground truth for this analysis. The closer to the ground truth a method is, the better its polarity prediction. SentiStrength was removed from this analysis as it was trained with this dataset.

We can make several interesting observations. First, we clearly see that most methods present more positive values than the negative values, as we see few lines below the ground truth among all the datasets.  Second, we note that several methods obtained only positive values, independent of the dataset analyzed. For instance, although SenticNet had a very high coverage, it identifies the wrong polarity for predominantly negative datasets.

\begin{figure*}[htpb!]
  \centering {
    \includegraphics[width=0.9\textwidth]{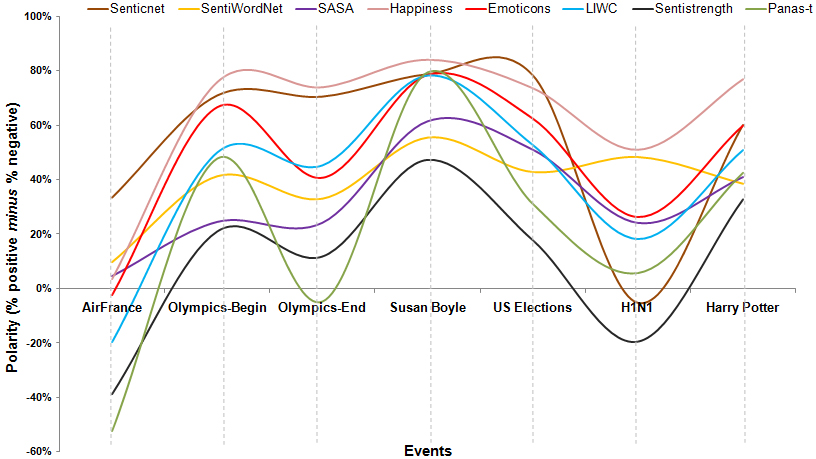}
  }
  \caption{Polarity of the eight sentiment methods across several real notable events,  indicating that existing methods vary widely in their agreement. }\label{fig:polarity-events}
\end{figure*}

This bias towards positive polarity showed by most of the methods might be trick for real time polarity detecting tools, as they might simply apply these methods in real time data, like Twitter streaming API, and account the rate of positive and negative message text. This would potentially show biased results due to the methods used. In order to verify this potential bias, we provide the same kind of analysis for each event we gathered from Twitter. Figure~\ref{fig:polarity-events} shows this polarity analysis. We can see that most of the methods show very positive results, even for datasets like H1N1. While this event's data may contain jokes and positive tweets, it would be also reasonable to expect a large number of tweets expressing concerns and bad feelings. Even the event related to an airplane crash was considered positive by four methods, although the polarity difference is close to zero for most of them.

\begin{figure*}[t!]
  \centering {
    \subfigure[Comparison]{\includegraphics[width=0.49\textwidth]{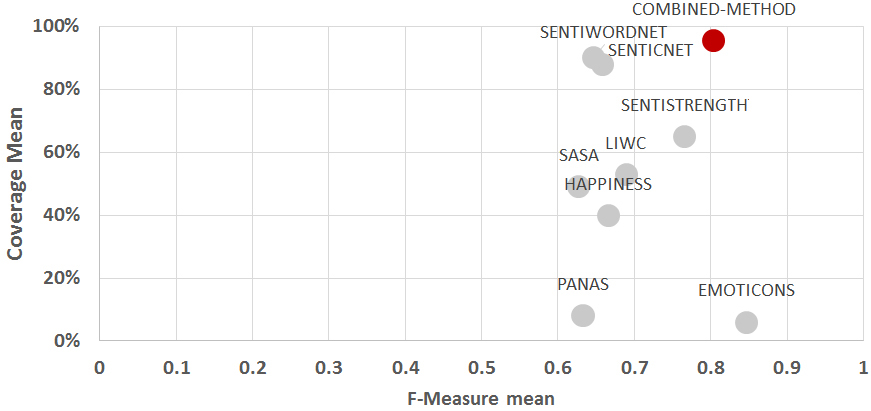}\label{fig:meta-method}}	 \subfigure[Tradeoff]{\includegraphics[width=0.49\textwidth]{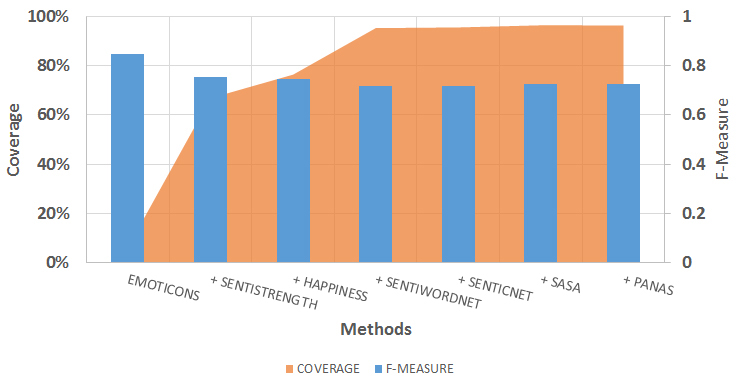}\label{fig:incremental_method_combination}}
}
  \caption{Trade off between the coverage vs F-measure for methods, including the proposed method}
\end{figure*}

\section{Combined method}
\label{sec:combining}

Having seen the varying degrees of coverage and agreement of the eight sentiment analysis methods, we next present a combined sentiment method and the comparison framework.

\subsection{Combined-Method}

We have created a combined method, which we simply call \textbf{Combined-method}. This new method relies on the following existing ones: PANAS-t, Emoticons, SentiStrength, SentiWordNet, SenticNet, SASA and Happiness Index. We omit LIWC due to copyright restrictions. The Combined-method analyzes the harmonic mean of the precision and recall for all methods and gives different weights for them (i.e., between 1 and 7). The goal is first to achieve the highest coverage, and then to achieve good agreement for a given dataset.

For evaluation, we tested the Combined-method over the SentiStrength labeled datasets~\cite{sentistrength1} that consist of human-labeled Web content (through AMT) drawn from Twitter, MySpace, Runners World, BBC, Digg, and YouTube (see description in Section~3.1.2). We calculated the average F-measure and the average coverage across these Web datasets. We also computed the coverage based on the near-complete Twitter dataset, averaging results over the six notable events  (see description in Section~3.1.1).

Figure~\ref{fig:meta-method} compares the coverage and the F-measure of the seven existing sentiment methods as well as the newly proposed Combined-method. The figure demonstrates the efficacy of Combined-method, in that it can detect sentiments with the highest coverage of 95\%, as one might expect. Furthermore, its accuracy and precision in sentiment analysis remain relatively high, with a F-measure of 0.730. This is lower than the best performing method, Emoticons, but higher than all other the other sentiment methods.

While combining all sentiment methods would yield the best coverage, there is a diminishing return effect, in that increasing the number of methods incurs only marginal gain in coverage after some point. Figure~\ref{fig:incremental_method_combination} shows this trend, where we add methods in the order of Emoticons, SentiStrength, Happiness Index, and so on (as noted in the horizontal axis). While Emoticons gives the lowest coverage of less than 10\%, the coverage increases to 70\% when we add just one more method, SentiStrength (see orange shaded region in the figure). The F-measure, on the other hand, drops slightly as more sentiment methods are combined, as seen in the blue-colored bars in the figure.

As we combine more methods, the coverage increases but to a smaller extent. In fact, combining the first four methods already achieves a coverage of 95\%, leaving only a small room for improvement after this point. We can also note that, although the accuracy and precision decrease as more methods are combined, they remain in a reasonable range (an F-measure of above 0.7). This indicates that combining all of the methods is not necessarily the best strategy. The best coverage and agreement may be achieved by combining those methods best suited for a particular kind of data. For example, one might want to choose LIWC over SASA for a given data or vice versa. Reducing the amount of data needed for Combined-method to obtain good results is a desirable property for a real system deployment, given that the use of fewer methods will likely require fewer resources.

\subsection{The iFeel Web System}

Finally, having compared the different sentiment methods and tested the efficacy of the Combined-method, we present for the research community a Web service called iFeel. iFeel
allows anyone on the Web to test the various sentiment analysis  methods compared in this paper with the texts of their choice. We exclude LIWC in the set of available tools,
due to copyright issues. The iFeel system was developed using Pyramid, an open source Web framework in Python based on Web Server Gateway Interface (WSGI).  A beta version of
the tool is available at \url{http://www.ifeel.dcc.ufmg.br}, and accepts short texts up to 200 characters as input.

\begin{figure}[!htpb]
	\centering
	\includegraphics[width=0.49\textwidth]{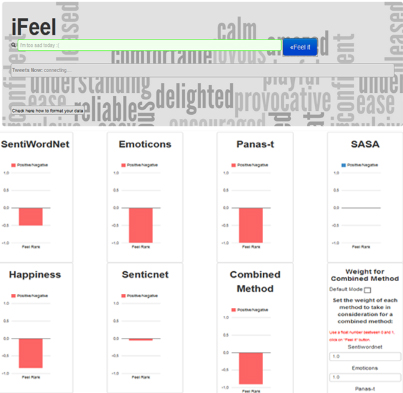}
	\caption{Screen snapshot of the iFeel web system}
	\label{fig:ifeel}
\end{figure}

Figure~\ref{fig:ifeel} shows the screen snapshot of the iFeel system for the input ``I'm feeling too sad today :(".  As demonstrated in this example, certain sentiment methods detect stronger degree of sentiment than others. For instance, Emoticons and PANAS-t can detect a high level of negative affect in this text, yet SenticNet and SASA do not.

In the future, we plan to extend the iFeel system to allow input files instead of input text strings, as well as supporting visualizations of different combinations of the
compared results. This will not only assist researchers in reproducing the experimental results presented in this paper and other papers, but also help users to decide on the
proper sentiment method for a particular dataset and application. Therefore, we hope that this system will be an important step towards applying sentiment analysis to OSN data.

\section{Concluding Remarks}

Recent efforts to analyze the moods embedded in Web 2.0 content have adapted various sentiment analysis methods originally developed in linguistics and psychology. Several of these methods became widely used in their knowledge fields and have now been applied as tools to measure polarity in the context of OSNs. In this paper, we present an apple-to-apple comparison of eight representative sentiment methods: SentiWordNet, SASA, PANAS-t, Emoticons, SentiStrength, LIWC, SenticNet, and Happiness Index.

Our comparison study focused on detecting the polarity of content (i.e., positive and negative affects) and does not yet consider other types of sentiments (e.g., psychological processes such as anger or calmness). We adopted two measures of efficacy, coverage (measuring the fraction of messages whose sentiments are detected) and agreement (measuring the fraction of identified sentiments that are in tune with ground truth). We find that the eight methods have varying degrees of coverage and agreement; no single method is always best across different text sources. This led us to combine the different methods to achieve the highest coverage and satisfying agreement; we presente this tool as the Combined-method.

We also present a Web API framework, called the iFeel system, through which we would like to allow other researchers to easily compare the results of a wide set of sentiment analysis tools. The system also gives access to the Combined-method, which typically gives the highest coverage and competitive accuracy. Although preliminary, we believe this is an important step toward a wider application of sentiment analysis methods to OSN data, able to help researchers decide on the proper sentiment method for a particular dataset and application.

This work has demonstrated a framework with which various sentiment analysis methods can be compared in an apple-to-apple fashion. To be able to do this, we have covered a wide range of research on sentiment analysis and have made significant efforts to contact the authors of previous works to get access to their sentiment analysis tools. Unfortunately, in many  cases, getting access to the tools was a nontrivial task; in this paper, we were only able to compare eight of the most widely used methods. As a natural extension of this work, we would like to continue to add more existing methods for comparison, such as the Profile of Mood States (POMS)~\cite{Bollen} and OpinionFinder~\cite{OpinionFinder}. Furthermore, we would like to expand the way we compare these methods by considering diverse categories of sentiments beyond positive and negative polarity.



\section{Acknowledgments}

The authors would like to thank the anonymous reviewers for their valuable comments. This work was funded by the Brazilian National Institute of Science and Technology for the Web (MCT/CNPq/INCT grant number 573871/2008-6) and grants from CNPq, CAPES and FAPEMIG. This paper was also funded by the Basic Science Research Program through the National Research Foundation of Korea funded by the Ministry of Education, Science and Technology (Project No. 2011-0012988) and the IT R\&D program of MSIP/KEIT (10045459, Development of Social Storyboard Technology for Highly Satisfactory Cultural and Tourist Contents based on  Unstructured Value Data Spidering).


\end{document}